\documentclass[11pt,a4paper]{article}
\usepackage[hyperref]{acl2020}
\usepackage{times}
\usepackage{latexsym}

\usepackage{microtype}
\usepackage{booktabs}
\aclfinalcopy % Uncomment this line for the final submission
\title{Self-Attention Gazetteer Embeddings for \\ Named-Entity Recognition}

\author{
  Stanislav Peshterliev, Christophe Dupuy, Imre Kiss \\
  Amazon, Alexa Machine Learning, USA \\
  \texttt{\{stanislp,dupuychr,ikiss\}@amazon.com}
 \\}

\date{}

\newcommand{\ourmethod}{GazSelfAttn}

\usepackage{numprint}
\usepackage{graphicx}
\usepackage{multirow}
\usepackage{textcomp}
\usepackage{amsmath}
\usepackage{enumitem}
\usepackage{nidanfloat}
\usepackage{etoolbox}

\usepackage[disable]{todonotes}

\def\*#1{\mathbf{#1}}

\begin{document}
\maketitle
\begin{abstract}
Recent attempts to ingest external knowledge into neural models for named-entity recognition (NER) have exhibited mixed results. In this work, we present~\ourmethod, a novel gazetteer embedding approach that uses self-attention and match span encoding to build enhanced gazetteer embeddings. In addition, we demonstrate how to build gazetteer resources from the open source Wikidata knowledge base. Evaluations on CoNLL-03 and Ontonotes 5 datasets, show F$_1$ improvements over baseline model from 92.34 to 92.86 and 89.11 to 89.32 respectively, achieving performance comparable to large state-of-the-art models.

\end{abstract}

\section{Introduction} 

Named-entity recognition (NER) is the task of tagging relevant entities such as person, location and organization in unstructured text. Modern NER has been dominated by neural models~\citep{lample2016neural,ma2016end} combined with contextual embeddings from language models (LMs)~\citep{peters2018deep, devlin2018bert, akbik2019pooled}. The LMs are pre-trained on large amounts of unlabeled text which allows the NER model to use the syntactic and semantic information captured by the LM embeddings. On the popular benchmark datasets CoNLL-03~\citep{sang2003introduction} and Ontonotes 5~\citep{weischedel2013ontonotes}, neural models with LMs achieved state-of-the-art results without gazetteers features, unlike earlier approaches that heavily relied on them~\citep{florian2003named}. Gazetteers are lists that contain entities such as cities, countries, and person names. The gazetteers are matched against unstructured text to provide additional features to the model. Data for building gazetteers is available for multiple language from structured data resources such as Wikipedia,  DBpedia~\citep{auer2007dbpedia} and Wikidata~\citep{vrandevcic2014wikidata}.

In this paper, we propose~\ourmethod, a novel gazetteer embedding approach that uses self-attention and match span encoding to build enhanced gazetteer representation.~\ourmethod~embeddings are concatenated with the input to a LSTM~\citep{hochreiter1997long} or CNN~\citep{strubell2017fast} sequence layer and are trained end-to-end with the model. In addition, we show how to extract general gazetteers from the Wikidata, a structured knowledge-base which is part of the Wikipedia project.  

Our contributions are the following:
\begin{itemize}[topsep=1pt, leftmargin=15pt, itemsep=-1pt]
\item We propose novel gazetteer embeddings that use self-attention combined with match span encoding.
\item We enhance gazetteer matching with multi-token and single-token matches in the same representation.
\item We demonstrate how to use Wikidata with entity popularity filtering as a resource for building gazetteers.
\end{itemize}

\ourmethod~evaluations on CoNLL-03 and Ontonotes 5 datasets show F$_1$ score improvement over baseline model from 92.34 to 92.86 and from 89.11 to 89.32 respectively. Moreover, we perform ablation experiments to study the contribution of the different model components.

\section{Related Work}

Recently, researchers added gazetteers to neural sequence models.~\citet{magnolini2019use} demonstrated small improvements on large datasets and bigger improvements on small datasets.~\citet{lin-etal-2019-gazetteer} proposed to train a gazetteer attentive network to learn name regularities and spans of NER entities. ~\citet{liu2019towards} demonstrated that trained gazetteers scoring models combined with hybrid semi-Markov conditional random field (HSCRF) layer improve overall performance. The HSCRF layer predicts a set of candidate spans that are rescored using a gazetteer classifier model. The HSCRF approach differs from the common approach of including gazetteers as an embedding in the model. Unlike the work of~\citet{liu2019towards}, our \ourmethod~does not require training a separate gazetteer classifier and the HSCRF layer, thus our approach works with any standard output layer such as conditional random field (CRF) \citep{lafferty2001conditional}.

\citet{wu2018evaluating} proposed an auto-encoding loss with hand-crafted features, including gazetteers, to improve accuracy. However, they did not find that gazetteer features significantly improve accuracy.

Extracting gazetteers from structure knowledge sources was investigated by \citet{torisawa2007exploiting} and \citet{ratinov2009design}. They used Wikipedia's \textit{instance of} relationship as a resource for building gazetteers with classical machine learning models. Compared to Wikidata, the data extracted from Wikipedia is smaller and noisier.

Similar to this paper, \citet{song2020improving} used Wikidata as a gazetteer resource. However, they did not use entity popularity to filter ambiguous entities and their gazetteer model features use simple one-hot encoding.

\section{Approach}

\subsection{Model Architecture}

\begin{figure}[!ht]
\includegraphics[width=7.5cm]{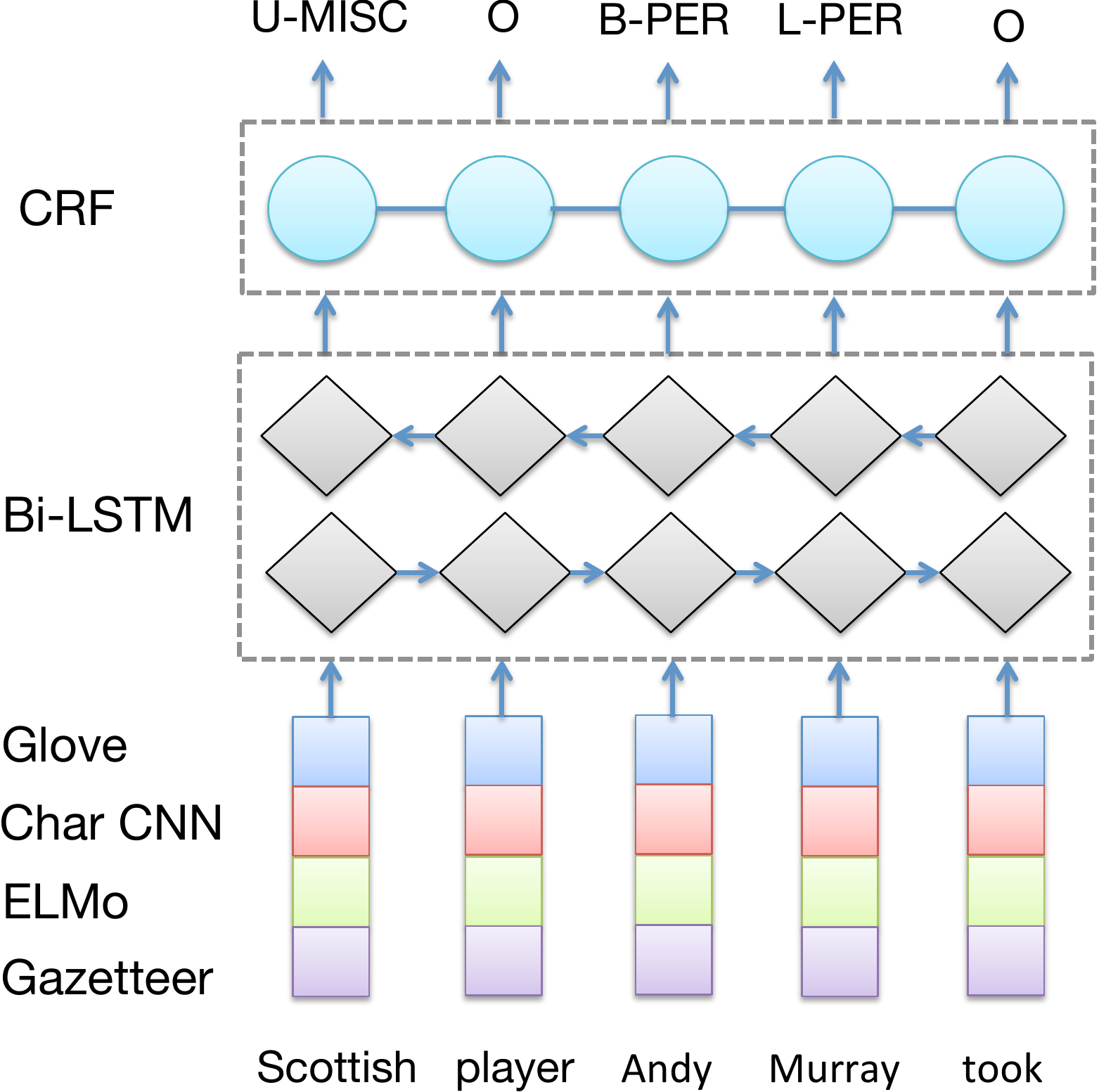}
\caption{Model architecture with gazetteer embeddings.}
\centering
\label{fig:arch}
\end{figure}

We add \ourmethod~embeddings to the popular Neural CRF model architecture with ELMo LM embeddings from~\citet{peters2018deep}. Figure~\ref{fig:arch} depicts the model, which consists of Glove word embeddings~\citep{pennington2014glove}, Char-CNN~\citep{chiu2016named, ma2016end}, ELMo embeddings, Bi-LSTM, and output CRF layer with BILOU (Beginning Inside Last Outside Unit) labels encoding~\citep{konkol2015segment}. Note that, we concatenate the gazetteer embeddings to the Bi-LSTM input.

\subsection{Gazetteers}

In this section, we address the issue of building a high-quality gazetteer dictionary $M$ that maps entities to types, e.g., $M$[Andy Murray] $\rightarrow$ Person. In this work, we use Wikidata, an open source structured knowledge-base, as the source of gazetteers. Although, Wikidata and DBpedia are similar knowledge bases, we choose Wikidata because, as of 2019, it provides data on around 45 million entities compared to around 5 million in DBpedia.

Wikidata is organized as entities and properties. Entities can be concrete (Boston, NATO, Michael Jordan) and abstract (City, Organization, Person). Properties describe an entity relations. For example, Boston \textit{instance\_of} City and Boston \textit{part\_of} Massachusetts; both instance\_of and part\_of are properties. Also, each entity is associated with sitelink count\footnote{\url{https://www.wikidata.org/wiki/Help:Sitelinks}} which tacks mentions of the entity on Wikimedia website and can be used as proxy for popularity.

To extract gazetteers from Wikidata, we process the official dumps\footnote{\url{dumps.wikimedia.org/wikidatawiki}} into tuples of entity and type based only on the left and right part of the instance\_of triplet, example resulting tuples are Boston $\rightarrow$ City and Massachusetts $\rightarrow$ State. Each entity is associated with a set of aliases, we keep only the aliases that are less than seven tokens long. Example aliases for Boston are ``Beantown'' and ``The Cradle of Liberty''. If there are multiple types per alias, we use the sitelink count to keep the six most popular types. The sitelink filtering is important to reduce the infrequent meanings of an entity in the gazetteer data.

The Wikidata types that we obtain after processing the Wikidata dumps are fine-grained. However, certain NER tasks require coarse-grained types. For instance, CoNLL-03 task has a single Location label that consists of cities, states, countries, and other geographic location. To move from fine-grained to coarse-grained types, we use the Wikidata hierarchical structure induced by the \textit{subclass\_of} property. Examples of subclass\_of hierarchies in Wikidata are: City $\rightarrow$ Human Settlement $\rightarrow$ Geographic Location, and Artist $\rightarrow$ Creator $\rightarrow$ Person. We change the types granularity depending on the NER task by traversing up, from fine-grained types to the target coarse-grained types. For instance, we merge  the Artist and Painter types to Person, and the River and Mountain types to Location.

\subsection{Gazetteer Matching}

Gazetteer matching is the process of assigning gazetteer features to sentence tokens. Formally, given a gazetteer dictionary $M$ that maps entities to types, and a sentence $S = (t_1, t_2, ..., t_n)$ with tokens $t_i$, we have to find the $m$ gazetteer types $\{g^1_i, g^2_i,..,g^m_i\}$ and spans $\{s^1_i, s^2_i,..,s^m_i\}$ for every token $t_i$. The set notation $\{$\} indicates that multiple $m$ matches are allowed per token. The match span $\{s^j_i\}$ represents positional information which encodes multi-token matches. The match spans are encoded using a BILU (Beginning Inside Last Unit) tags, similar to the BILOU tags that we use to encode the NER labels.

In general, there are two methods for gazetteer matching: multi-token and single-token. \textit{Multi-token} matching is searching for the longest segments of the sentence that are in $M$. For instance, given $M$[New York] $\rightarrow$ State, $M$[New York City] $\rightarrow$ City and the sentence ``Yesterday in New York City'', the multi-token matcher assigns the City gazetteer type to the longest segment ``New York City''. \textit{Single-token} matching is searching to match any vocabulary word from a gazetteer type. In the earlier example, each word from the sentence is individually matched to the tokens in $M$, thus ``New'' and ``York'' are individually matched to both City and State, and ``City'' is matched only to City. 

\citet{magnolini2019use} research shows that both multi-token and single-token matching perform better on certain datasets. We propose to combine both methods: we tag the multi-token matches with BILU tags, and the single-token matches with a Single (S) tag. The single-token matches are used only if multi-token matches are not present. We consider that the single-token matches are high-recall low-precision, and multi-token matches are low-recall and high-precision. Thus, a combination of both works better than individually. Example sentences are: ``Yesterday in New(City-B) York(City-I) City(City-L)'', and ``Yesterday in York(City-S) City(City-S)'' York City is marked with singles tag since $M$ does not have entities for ``York City'', ``York'', and ``City''.

Note that gazetteer matching is unsupervised, i.e., we do not have a ground truth of correctly matched sentences for $M$. Furthermore, it is a hard task because of the many variations in writing and ambiguity of entities.

\subsection{Gazetteer Embeddings}

\begin{align}
\*E_i &= \*G[\{g^m_i\}] \oplus \*S[\{s^m_i\}] \label{embed} \\ 
\*A_i &= \textrm{softmax}(\frac{\*E_i \*E_i^T}{\sqrt{d}})\*E_i \label{attn} \\ 
\*H_i &= \textrm{GELU}(\*W \*A_i + \*{b}) \label{feed} \\ 
\*{g}_i &= \textrm{maxpooling}(\*H_i) \label{pooling}
\end{align}
\vspace{-15px}

Equations~\ref{embed}-\ref{pooling} shows the gazetteer embedding $\*g_i$ computation for a token $t_i$. To compute $\*g_i$, given a set of $m$ gazetteer types $\{g^m_i\}$ and spans $\{s^m_i\}$, we execute the following procedure:
\begin{itemize}[topsep=1pt, leftmargin=15pt, itemsep=-1pt]
\item Equation~\ref{embed}. We embed the sets $\{g^m_i\}$ and $\{s^m_i\}$ using the embedding matrices $\*G$ and $\*S$. Then, we do an element-wise addition, denoted $\oplus$, of the corresponding types and spans embeddings to get a matrix $\*E_i$.
\item Equation~\ref{attn}. We compute $\*A_i$ using scaled dot-product self-attention~\citep{attnallyouneed}, where $d$ is the dimensionality of the gazetteer embeddings. The attention contextualizes the embeddings with multiple gazetteer matches per token $t_i$.
\item Equation~\ref{feed}. To add model flexibility, we compute $\*H_i$ with a position-wise feed-forward layer and GELU activation~\citep{hendrycks2016gaussian}.
\item Equation~\ref{pooling}. Finally, we perform max pooling across the embeddings $\*H_i$ to obtain the final gazetteer embedding $\*g_i$.
\end{itemize}

\subsection{Gazetteer Dropout}

To prevent the neural NER model from overfitting on the gazetteers, we use gazetteers dropout~\citep{yang2016drop}. We randomly set to zero gazetteer embeddings $\*g_i$, so the gazetteer vectors that are input to the LSTM become zero. We tune the gazetteer dropout hyperparameter on the validation set.

\section{Experiments}
\subsection{Setup}
\textbf{Datasets.} We evaluate on the English language versions of CoNLL-03 dataset \citep{sang2003introduction} and the human annotated portion of the Ontonotes 5 \citep{weischedel2013ontonotes} dataset. CoNLL-03 labels cover 4 entity types: person, location, organization, and miscellaneous. The Onotonotes 5 dataset is larger and its labels cover 18 types: person, NORP, facility, organization, GPE, location, product, event, work of art, law, language, date, time, percent, money, quantity, ordinal, cardinal.

\begin{table}[h!]
\centering
\begin{center}
\begin{tabular}{ c|c c c } 
 \hline
 \textbf{Dataset} & \textbf{Train} & \textbf{Dev} & \textbf{Test} \\  \hline
 CoNLL-03 & \numprint{14987} & \numprint{3466} & \numprint{3684} \\  \hline
 Onotonotes 5 & \numprint{82728} & \numprint{10508} & \numprint{10394} \\  \hline
\end{tabular}
\end{center}
\caption{Dataset sizes in number of sentences.}
\label{table:datasets}
\end{table}
\vspace{-10px}

\textbf{Gazetteers.} We use the Wikidata gazetteers with types merged to the granularity of the CoNLL-03 and Ononotes 5 datasets. We filter non-relevant types (e.g., genome names, disease) and get a total of one million records. For CoNLL-03 and Ontonotes 5, the percentage of entities covered by gazetteers are 96\% and 78\% respectively, and percentage of gazetteers wrongly assigned to non-entity tokens are 41\% and 41.5\% respectively.

\textbf{Evaluation.} We use the standard CoNLL evaluation script which reports entity F1 scores. The F1 scores are averages over 5 runs.

\textbf{Configuration.} We use the Bi-LSTM-CNN-CRF model architecture with ELMo language model embeddings from~\citet{peters2018deep}, which consist of 50 dim pre-trained Glove word embeddings~\citep{pennington2014glove}, 128 dim Char-CNN~\citep{chiu2016named, ma2016end} embeddings with filter size of 3 and randomly initialized 16 dim char embeddings, 1024 pre-trained ELMo pre-trained embeddings, two layer 200 dim Bi-LSTM, and output CRF layer with BILOU (Beginning Inside Last Outside Unit) spans~\citep{konkol2015segment}.

For the gazetteer embeddings, we use 128 dim for the embedding matrices $\*G$ and $\*S$, 128 dim output for $\*W$, which yields a gazetteer embedding $\*g_i$ with 128 dim. The parameters are randomly initialized and trained. We apply gazetteer dropout of 0.1 which we tuned on the development set; we tried values form 0.05 to 0.6.

All parameters except the ELMo embeddings are trained. We train using the Adam~\citep{kingma2014adam} optimizer with learning rate of 0.001 for 100 epochs. We use early stopping with patience 25 on the development set. Batch size of 64, dropout rate of 0.5 and L2 regularization of 0.1.

\subsection{Results}

\begin{table}[!h]
\centering
\begin{center}
\begin{tabular}{ l|cc } 
 \hline
 \multirow{2}{4em}{\textbf{Model}} & \multicolumn{2}{c}{\textbf{Test F$_1$\textpm~std}} \\ 
  & CoNLL-03 & Ontonotes 5 \\ \hline
 \citet{devlin2018bert} & 92.8 & - \\ 
 \citet{peters2018deep} & 92.22\textpm0.10 & 89.13\textpm0.23 \\
 \citet{akbik2019pooled} & 93.09\textpm0.12 & 89.3 \\ 
 \citet{baevski2019cloze} & 93.5 & - \\
 \citet{liu2019towards} & 92.75\textpm0.18 & - \\ \hline
 ELMo Neural CRF & 92.34\textpm0.12 & 89.11\textpm0.23 \\ 
 + Our~\ourmethod & 92.86\textpm0.13 & 89.32\textpm0.21 \\
 Neural CRF & 90.42\textpm0.10 & 86.63\textpm0.18 \\ 
 + Our~\ourmethod & 91.12\textpm0.12 & 86.87\textpm0.21 \\ \hline
\end{tabular}
\end{center}
\caption{Results on CoNLL-03 and OntoNotes 5.}
\label{table:ner_datasets}
\end{table}

The experimental results for NER are summarized in Table~\ref{table:ner_datasets}. The top part of the table shows recently published results. \citet{liu2019towards}'s work is using gazetteers with HSCRF and \citet{akbik2019pooled}'s work is using the Flair language model which is much larger than ELMo. \citet{baevski2019cloze} is the current state-of-the-art language model that uses cloze-driven pretraining. The bottom part of the table is shows our baseline models and results with included gazetteers. We experiment with the Neural CRF model with and without ELMo embeddings. Including ELMo embeddings the CoNLL-03 and Ontonotes 5, F$_1$ score improves from 92.34 to 92.86 and 89.11 to 89.32 respectively. Without ELMo embeddings the F$_1$ score improves from 90.42 to 91.12 and 86.63 to 87 respectively. We observe that \ourmethod~relative improvements are similar with and without ELMo embeddings. We obtain slightly better CoNLL-03 F$_1$ score compared to \citet{liu2019towards} work that uses the HSCRF model, and we match the Ononotes 5 F$_1$ scores of \citet{akbik2019pooled} that uses a much bigger model. \citet{liu2019towards} Ononotes 5 results use subset of the dataset labels and are not comparable. Note that because of computation constrains, we did not perform extensive hyperparameter tuning except for the gazetteer dropout rate.

\subsection{Ablation study}

\begin{table*}[!t]
\centering
\begin{tabular}{ l|cc cc } 
 \hline
 \multirow{2}{4em}{\textbf{Model}} & \multicolumn{2}{c}{\textbf{CoNLL-03}} & \multicolumn{2}{c}{\textbf{Ontonotes 5}} \\ 
  & Dev F$_1$\textpm~std & Test F$_1$\textpm~std & Dev F$_1$\textpm~std & Test F$_1$\textpm~std \\ \hline
% ELMo-LSTM-CRF + GazEmbed & - & - & - & -  \\ 
% - span & - & - & - & -  \\ 
% - self attention + max pooling & - & - & - & -  \\ 
% - single matches & - & - & - & -  \\ 
% - multi matches & - & - & - & -  \\ 
 Neural CRF + \ourmethod & 95.01\textpm0.11 & 91.1\textpm0.12 & 85.05\textpm0.20 & 86.87\textpm0.21  \\ 
 - span encoding & 94.86\textpm0.11 & 90.82\textpm0.14 & 85.25\textpm0.20 & 86.81\textpm0.20  \\ 
 - self-attention & 94.90\textpm0.11 & 90.61\textpm0.11 & 85.05\textpm0.11 & 86.71\textpm0.11  \\
 - single matches & 94.85\textpm0.11 & 90.55\textpm0.11 & 84.95\textpm0.11 & 86.69\textpm0.11  \\  
% - multi matches & - & - & - & -  \\  
\end{tabular}
\caption{Ablation study results on CoNLL-03 and OntoNotes 5. ``- span encoding'' removes the BILU match span encoding leaving only the gazetteer types. ``- self attention'' removes the self-attention. ``- uncased matches'' removes the uncased matches.}
\label{table:ablation}
\end{table*}

Table~\ref{table:ablation} shows ablation experiments. We remove components of the gazetteer embedding model from the Neural CRF model. In each experiment, we removed only the specified component. Ablations show decreased F$_1$ score on the development and test set if any of the components is removed. The highest degradation is when single matches are removed which underscores the importance of the combining the gazetteer matching techniques for NER. We observe that match span encoding is more important for the CoNLL-02 compared to Ononotes 5 because the former has more entities with multiple tokens. Removing the self-attention shows that self-attention is effective at combining information form multiple gazetteers.

In addition, we tried moving the gazetteer embeddings to the CRF layer and using the LSTM token embeddings as attention keys but the F$_1$ degraded significantly. We experimented with adding auto-encoding loss similar to \citet{wu2018evaluating} and multi-head self-attention. However, we did not observe F$_1$ score improvements and sometimes small degradations.

\section{Conclusion}

We presented \ourmethod, a novel approach for gazetteer embeddings that uses self-attention and match span positions. Evaluation results of \ourmethod~show improvement compared to competitive baselines and state-of-the-art models on multiple datasets.

For future work we would like to evaluate \ourmethod~on non-English language datasets and improve the multi-token gazetteer matching with fuzzy string matching. Also, we would like to explore transfer learning of gazetteer embeddings from high-resource to low-resource setting.

\bibliography{acl2020}
\bibliographystyle{acl_natbib}

\end{document}